\documentclass[letterpaper, 10 pt, conference]{ieeeconf}  

\IEEEoverridecommandlockouts                              

\usepackage{graphics} 
\usepackage{epsfig} 
\usepackage{mathptmx} 
\usepackage{times} 
\usepackage{amsmath} 
\usepackage{amssymb}  
\usepackage{subcaption}
\usepackage{cite}
\usepackage{makecell}
\usepackage{multirow}
\usepackage{caption}
\usepackage{supertabular}
\usepackage{float}
\usepackage{booktabs}
\usepackage{tabularx}
\usepackage{graphicx}
\usepackage{caption}
\usepackage{placeins}
\usepackage{subcaption} 

\makeatletter
\newif\if@restonecol
\makeatother

\usepackage[linesnumbered,ruled,vlined]{algorithm2e}
\usepackage{algpseudocode}
\usepackage{amsmath}

\title{\LARGE \bf
Breaking the Latency Barrier: Synergistic Perception and Control for High-Frequency 3D Ultrasound Servoing*
}

\title{\LARGE \bf
Breaking the Latency Barrier: Synergistic Perception and Control for High-Frequency 3D Ultrasound Servoing
}

 \author{
     Yizhao Qian,
     Yujie Zhu,
     Jiayuan Luo,
     Li Liu,
     Yixuan Yuan,
     Guochen Ning$^*$,
     Hongen Liao
     \thanks{Yizhao Qian and Yixuan Yuan are with the Department of Electronic Engineering, The Chinese University of Hong Kong, Hong Kong SAR, China.}
     \thanks{Guochen Ning, Hongen Liao, and Yujie Zhu are with Department of Biomedical Engineering, Tsinghua University, Beijing, China.}
     \thanks{Li Liu and Jiayuan Luo are with Great Bay University, Dongguan, China.}
     \thanks{$^*$Corresponding author: Guochen Ning (email: ningguochen@tsinghua.edu.cn).}
 }

\begin{document}
\bstctlcite{IEEEexample:BSTcontrol} 
\maketitle
\thispagestyle{empty}
\pagestyle{empty}

\begin{abstract}

Real-time tracking of dynamic targets amidst large-scale, high-frequency disturbances remains a critical unsolved challenge in Robotic Ultrasound Systems (RUSS), primarily due to the end-to-end latency of existing systems. This paper argues that breaking this latency barrier requires a fundamental shift towards the synergistic co-design of perception and control. We realize it in a novel framework with two tightly-coupled contributions: (1) a Decoupled Dual-Stream Perception Network that robustly estimates 3D translational state from 2D images at high frequency, and (2) a Single-Step Flow Policy that generates entire action sequences in one inference pass, bypassing the iterative bottleneck of conventional policies. This synergy enables a closed-loop control frequency exceeding $60~Hz$. On a dynamic phantom, our system not only tracks complex 3D trajectories with a mean error below $6.5~mm$ but also demonstrates robust re-acquisition from over $170~mm$ displacement. Furthermore, it can track targets at speeds of $102~mm/s$, achieving a terminal error below $1.7~mm$. Moreover, in-vivo experiments on a human volunteer validate the framework's effectiveness and robustness in a realistic clinical setting. Our work presents a RUSS holistically architected to unify high-bandwidth tracking with large-scale repositioning, a critical step towards robust autonomy in dynamic clinical environments.

\end{abstract}

\section{INTRODUCTION}

\begin{figure}
    \centering
    \setlength{\belowcaptionskip}{-0.8cm}
    \includegraphics[width=1\linewidth]{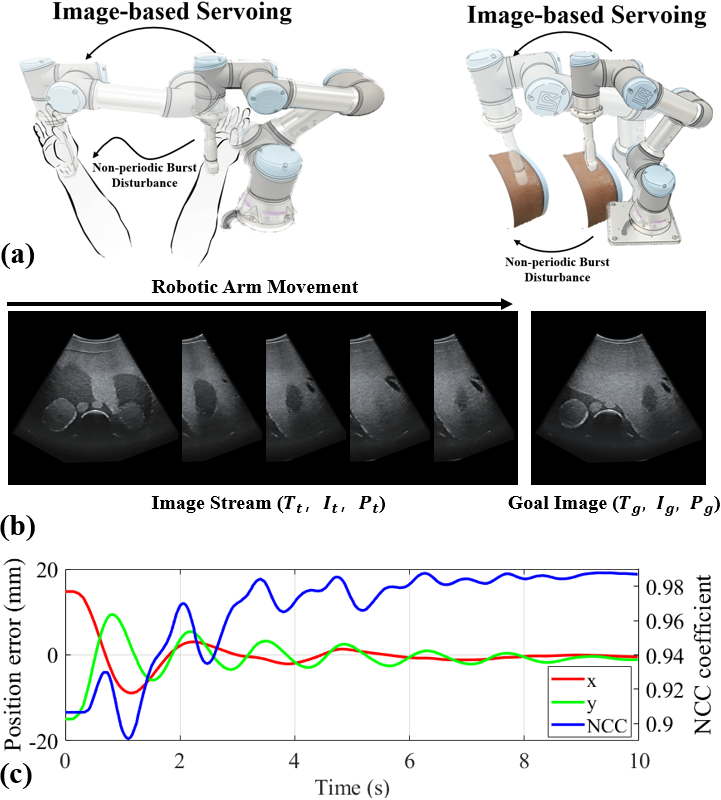}
    \caption{
    Overview of the proposed high-frequency visual servoing. 
    (a) Challenge: maintaining the ultrasound view under significant and unpredictable disturbances. 
    (b) Control objective: aligning the live video stream with the target image using robotic manipulation. 
    (c) Outcome: rapid reduction of positional errors (x, y) and maximization of image similarity, quantified by normalized cross-correlation (NCC).
    }
    \label{fig1}
\end{figure}

Robotic Ultrasound Systems (RUSS) hold immense potential to revolutionize medical diagnostics\cite{8,9}. The ultimate vision is a fully autonomous system capable of expert-level dexterity, yet this is critically hindered by a core challenge: tracking anatomical targets amidst high-frequency, unpredictable physiological motions. As illustrated in Fig.~\ref{fig1}, the fundamental task is to dynamically adjust the robotic arm to minimize the error between a live image stream and a goal image, maintaining high-quality visualization despite constant disturbances. However, the inherent latency in existing systems makes achieving this in real-time a significant and unresolved problem in robotic sonography.

Prevailing RUSS paradigms are ill-suited for this challenge: event-driven strategies\cite{2} are too slow for continuous motion, model-driven servoing\cite{10,4} suffers from long system-level convergence times, and even state-of-the-art learning methods like Diffusion Policy are bottlenecked by slow iterative inference\cite{30}, capping their control frequency below the \textbf{60 Hz frame rate typical of medical ultrasound (US) probes and image grabber}. Any control loop slower than this rate cannot respond to every new frame of information, making true real-time tracking unattainable. This exposes a clear technological gap: the absence of a holistic framework architected from the ground up for high-bandwidth, real-time tracking.

To address this critical gap, we propose a framework founded on the central principle of \textbf{synergistic co-design} for minimal latency. Our core insight is that a fast policy is ineffective without an equally fast perception module providing timely state information, and vice versa. This synergy is realized through two cornerstone innovations: our \textbf{Decoupled Dual-Stream Perception Network}, which uniquely separates in-plane geometric matching from out-of-plane semantic inference to robustly estimate 3D state at high frequency; and our \textbf{Single-Step Flow Policy}, which leverages a Flow model to generate an entire predictive action sequence in a single forward pass, fundamentally removing the iterative latency of prior generative models. This tightly-integrated perception-control loop is paired with a sample-efficient Sim-to-Real strategy, designed to leverage the decoupled nature of our perception front-end for rapid adaptation. The efficacy of our framework is validated not only on a dynamic phantom but also through an in-vivo study on a human volunteer. The main contributions of this paper are:

\begin{itemize}
\item A novel RUSS framework that synergistically integrates a high-frequency Flow Policy with a co-designed perception front-end, achieving unprecedented 62 Hz closed-loop tracking of dynamic targets.
\item A fast, dual-stream perception architecture that resolves the key ambiguity of out-of-plane motion estimation, enabling robust, real-time 3D translational servoing from 2D images.
\item A demonstration of sample-efficient Sim-to-Real transfer, where our framework generalizes from simulation to a physical phantom using only 50 expert trajectories with rapid convergence.
\end{itemize}





\section{RELATED WORK}

\subsection{Robotic Ultrasound Systems: The System-Level Bottleneck for Dynamic Tracking}

Recent advances in RUSS have shown success in automating quasi-static tasks like vascular screening \cite{10}, thyroid scanning \cite{20}, and standard plane localization \cite{17,4,19}. The feasibility of maintaining stable probe contact is also well-established \cite{5}.

However, addressing patient and target motion, particularly high-frequency, unpredictable disturbances, remains a formidable challenge. One common paradigm, event-driven discrete compensation, employs a "Stop-Register-Resume" strategy\cite{2}, but its reported 336 ms registration latency makes it fundamentally untenable for continuous clinical disturbances.

Another line of work pursues continuous tracking via model-driven visual servoing, achieving high control frequencies\cite{3} (20 Hz) or sub-millimeter\cite{1,4} static accuracy. Yet, their system-level responsiveness is poor, with end-to-end convergence times on the order of seconds, even when using high-rate perception (60 Hz) \cite{4,10,1}. This discrepancy proves a critical point: \textbf{component-level speed does not translate to system-level agility}, exposing the need for a framework holistically architected for low-latency dynamic response.

Therefore, a critical gap exists for a RUSS framework holistically architected for high-bandwidth, unpredictable motion tracking. Recent surveys confirm that the lack of real-time \cite{8} integrated perception and control \cite{8,9} is a key challenge in the field. Our work directly addresses this gap by proposing a framework where these subsystems are synergistically co-designed for a low-latency dynamic response.





\begin{figure*}[htb]
    \centering
    \setlength{\belowcaptionskip}{-0.5cm}
    \includegraphics[width=1\linewidth]{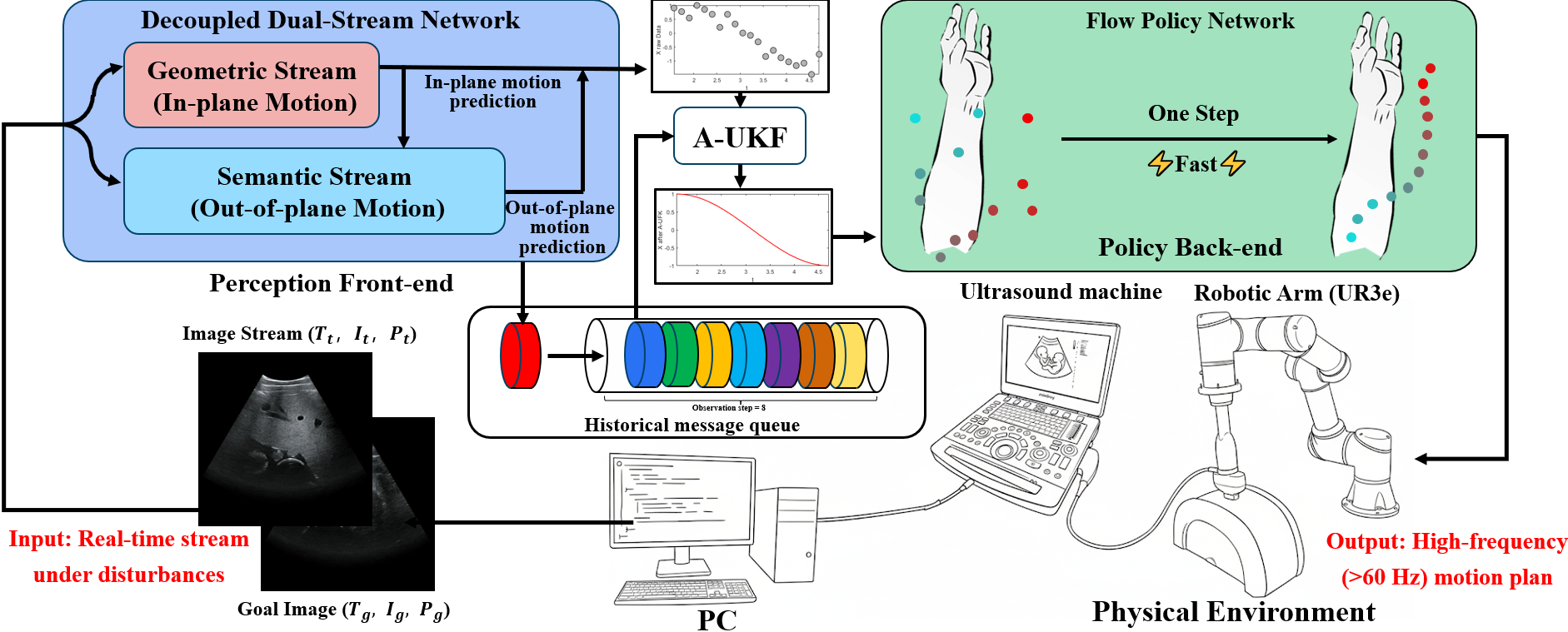}
    \caption{Overview of our proposed high-frequency US servoing framework. The system takes a live image stream and a goal image as input. The Vision Front-end, composed of a Decoupled Dual-Stream Network and an Adaptive-UKF, estimates the 3D translational error. This state information is fed to the Flow Policy Network, which generates a short-horizon motion plan executed by the robotic arm in the Physical Environment.}
    \label{fig:framework}
\end{figure*}

\subsection{Learning-based Control: The Quest for High-Frequency Policies and Robust Generalization}

Learning-based methods, particularly imitation learning (IL), are effective for acquiring expert workflows in RUSS \cite{ning1,43}. The state-of-the-art is dominated by Diffusion Policies \cite{27}, but their reliance on an iterative denoising process for inference imposes a fundamental latency bottleneck. This limits their control frequency to ~10-23 Hz \cite{27, 31}, a rate far below the ~60 Hz update stream from the US probe, making real-time compensation of physiological motion impossible.

To overcome this, policies based on Flow Matching have emerged as a compelling alternative \cite{28}. By enabling single-step inference, their recent work demonstrating speeds of ~50 Hz—a nearly 7-fold improvement \cite{29} over diffusion counterparts and highlighting their potential for high-frequency control\cite{30}.

However, a fast policy alone is insufficient. A critical second challenge is generalization against variations in US appearance. Existing frameworks are often too slow for dynamic tasks, relying on minute-long offline searches \cite{24} or using perception modules that limit the system frame rate to a mere 3 fps \cite{17}. This reveals a critical trade-off: existing methods sacrifice either real-time performance for generalization, or vice-versa.

Therefore, an effective framework must address both challenges in concert. To our knowledge, no prior work has presented a holistic framework where a high-frequency policy is synergistically co-designed with a fast, sample-efficient Sim-to-Real strategy to enable true, end-to-end dynamic tracking at over 60 Hz. This fusion of a high-bandwidth policy with a robust, low-latency generalization pipeline is the central methodological contribution of our paper.





\subsection{The Perception Bottleneck for High-Frequency Servoing}

The performance of any high-frequency control system is ultimately limited by the latency and accuracy of its perception front-end. Common RUSS perception pipelines, comprising segmentation, feature extraction, and matching, inherently accumulate latency and propagate errors \cite{4,13}, rendering them unsuitable for real-time dynamic tracking.

This challenge is particularly acute in US due to a fundamental ambiguity: inferring out-of-plane (Z) motion from a 2D image sequence is a notoriously ill-posed problem \cite{3,38}. Existing systems often circumvent this with inefficient search strategies or are confined to 2D in-plane compensation only \cite{6}. While end-to-end regression has been proposed \cite{11}, these methods have not been validated within a high-frequency dynamic tracking loop.

Therefore, a perception module for this task must be low-latency and architected to resolve the out-of-plane ambiguity from image data directly. We address this by proposing a novel, decoupled dual-stream architecture that estimates the full 3D translational state at high frequency. This perception front-end is co-designed with our high-speed policy, forming the cornerstone of our synergistic framework.





\section{METHODOLOGY}

\label{sec:methodology}

\subsection{Problem Formulation}
\label{sec:problem_formulation}
We formulate the dynamic visual servoing task as learning a policy, $\pi_{\theta}$, that maps a history of visual observations to a sequence of future actions. The goal is to minimize the 3D translational error, $\mathbf{e}_t$, between the live US stream, $I_t$, and a static goal image, $I_g$.

At each time step $t$, a perception front-end, $\phi$, estimates this error (detailed in Sec.~\ref{sec:perception_module}):
\begin{equation}
    \mathbf{e}_t = [dx_t, dy_t, dz_t]^T = \phi(I_t, I_g)
    \label{eq:perception}
\end{equation}
where $dx_t, dz_t$ are in-plane errors and $dy_t$ is the out-of-plane error. To capture target dynamics, the system state, $\mathbf{s}_t$, is defined as a temporal sequence of the $k$ most recent errors ($k=8$ in our work):
\begin{equation}
    \mathbf{s}_t = (\mathbf{e}_t, \mathbf{e}_{t-1}, \dots, \mathbf{e}_{t-k+1}) \in \mathbb{R}^{3 \times k}
    \label{eq:state}
\end{equation}
Unlike reactive approaches \cite{41}, our policy outputs a short-horizon motion plan of $H$ future actions ($H=8$), where each action $\mathbf{a}_{t+i} \in \mathbb{R}^3$ is a desired translational velocity command $[v_x, v_y, v_z]^T$:
\begin{equation}
    \mathbf{A}_t = (\mathbf{a}_t, \mathbf{a}_{t+1}, \dots, \mathbf{a}_{t+H-1})
    \label{eq:action_sequence}
\end{equation}
The core task is to learn the deterministic policy $\pi_{\theta}$ that maps the state history to this action sequence:
\begin{equation}
    \mathbf{a}_t, \dots, \mathbf{a}_{t+H-1} = \pi_\theta(\mathbf{s}_t)
    \label{eq:policy}
\end{equation}
During execution, we employ a receding horizon strategy, applying the first $h=4$ actions of the predicted sequence before the policy is re-evaluated.


\subsection{Framework Overview}
\label{sec:framework_overview}

Our solution to this predictive control problem is a novel framework for high-frequency dynamic visual servoing, illustrated in Fig.~\ref{fig:framework}. Our framework is architected for minimal end-to-end latency, adhering to the principle of synergistic co-design. This synergy is not merely about combining fast components, but about ensuring a seamless, high-bandwidth flow of information. The framework tightly integrates a high-frequency perception front-end (Sec.~\ref{sec:perception_module}) with a single-step predictive policy (Sec.~\ref{sec:policy_module}). This ensures that the policy's minimal inference latency is not wasted waiting for perception, and the perception's high-rate state estimates are immediately acted upon. The result is a complete perception-to-action loop operating at over $60$ Hz, enabling decisive compensation of dynamic disturbances.

\subsection{High-Frequency Temporal Perception Front-End}
\label{sec:perception_module}

The core task of our perception front-end is to robustly estimate the target's 3D translational motion from a 2D US stream. This requires balancing two conflicting objectives: \textbf{(1) Generalization} for performance across diverse subjects, and \textbf{(2) Real-time Performance} for tracking high-frequency motion at over 60 Hz. Our solution is a synergistic system composed of a structured visual observer and a predictive temporal filter.

\subsubsection{Decoupled Architecture for Generalizable 3D Motion Features}
\label{sec:decoupled_arch}

\begin{figure}
    \centering
    \setlength{\belowcaptionskip}{-0.5cm}
    \includegraphics[width=1\linewidth]{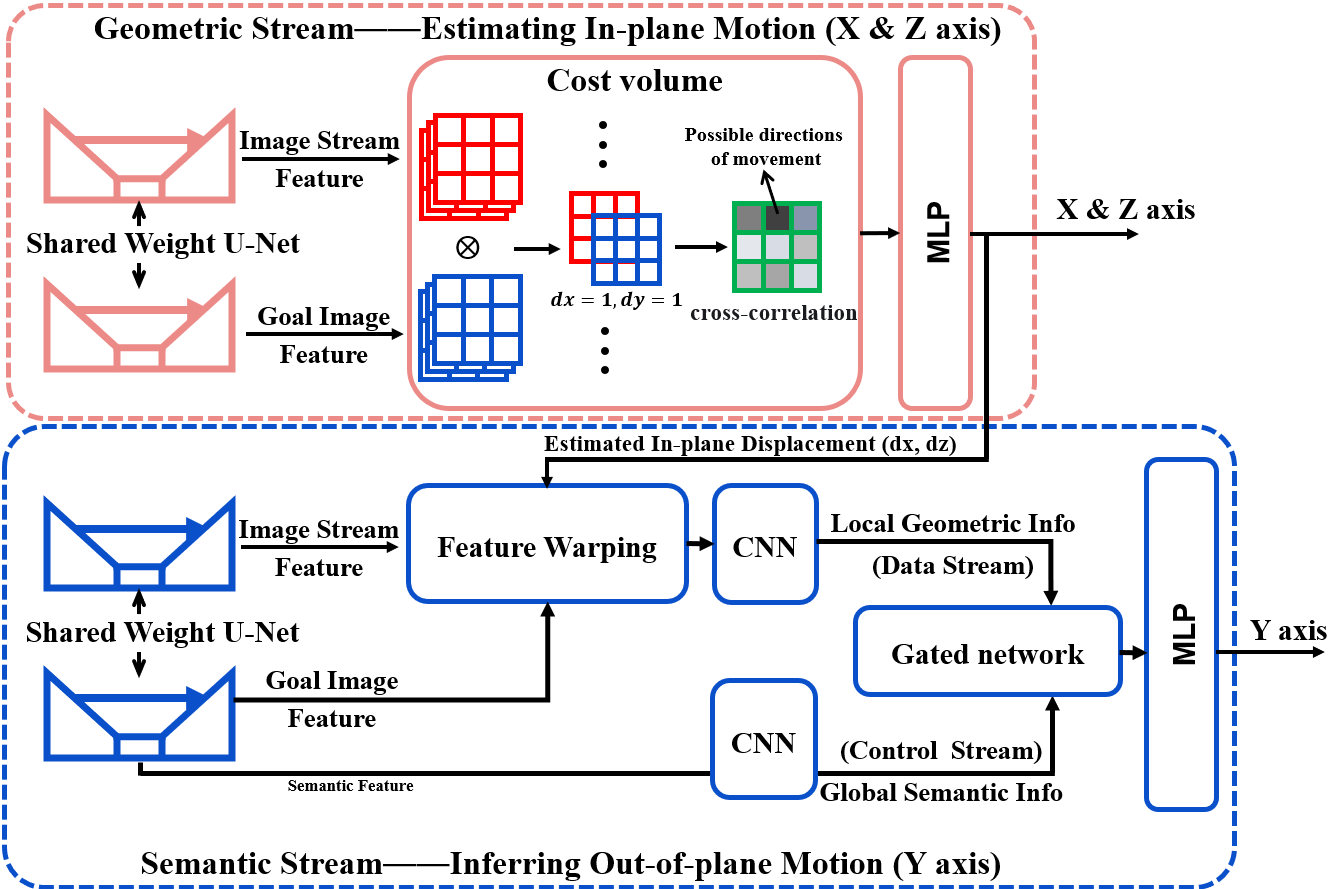}
    \caption{The architecture of our Decoupled Dual-Stream Perception Network. The Geometric Stream uses a cost volume to estimate in-plane motion (X \& Z axis) based on low-level feature. Concurrently, the Semantic Stream infers out-of-plane motion (Y axis) by interpreting higher-level feature.}
    \label{fig:dual_unet_diagram}
\end{figure}

To promote generalization, we avoid a "black-box" design and instead propose a structured architecture with physically-motivated inductive biases (Fig.~\ref{fig:dual_unet_diagram}). We decouple the estimation of in-plane and out-of-plane motion, as they stem from fundamentally different visual cues.

This architecture consists of two specialized, parallel streams. The \textbf{Geometric Stream} first estimates in-plane motion $(\mathbf{d_xz}=[d_x, d_z])$ by performing dense matching on low-level geometric feature maps, $\mathbf{\phi}_g(\cdot)$, using a cost volume:
\begin{equation}
    C(u, v, \mathbf{d}) = \langle \mathbf{\phi}_g(I_g)_{u,v}, \mathbf{\phi}_g(I_t)_{u+d_x, v+d_z} \rangle.
    \label{eq:cost_volume}
\end{equation}
This reliance on geometric correspondence makes it inherently robust to appearance shifts. Crucially, the estimated in-plane displacement $\mathbf{d}$ is then used to warp the feature maps for the second stream. The \textbf{Semantic Stream} analyzes these warped high-level semantic features, $\mathbf{\phi}_s(\cdot)$, to infer the more ambiguous out-of-plane motion ($d_y$). It is trained to interpret changes in anatomical morphology as translational displacement.

This architectural decoupling is key to our sample-efficient Sim-to-Real strategy. It allows for a targeted fine-tuning process where the domain-sensitive Semantic Stream is fully trained, while the Geometric Stream is largely frozen, with only the initial and final layers being fine-tuned to adapt to real-world texture and scaling variations.

\subsubsection{Predictive State Estimation for Real-Time Performance}
\label{sec:filtering}

To ensure real-time throughput, we employ computational optimizations such as feature caching and pre-computation for the static goal image. The raw 3D motion estimates, $\mathbf{d}_t = [d_x, d_y, d_z]^T$, from the vision network are inherently noisy. To address this, we use a filter to produce a smooth, predictive state estimate for the policy.

We chose an \textbf{Adaptive Unscented Kalman Filter (A-UKF)} over a standard Extended Kalman Filter (EKF) for its superior performance with potentially non-linear system dynamics without requiring the computation of Jacobians. The A-UKF's primary role is to smooth the noisy perception measurements and provide a rich, predictive state estimate $\mathbf{\hat{x}}_t$ for the policy. This state explicitly models not only the 3D position $\mathbf{p}_t$ and velocity $\mathbf{v}_t$, but also the critical sensor bias $\mathbf{b}_t$:
\begin{equation}
    \mathbf{x}_t = [\mathbf{p}_t^T, \mathbf{v}_t^T, \mathbf{b}_t^T]^T \in \mathbb{R}^9
\end{equation}
By fusing temporal information and accounting for system bias, the A-UKF provides the high-quality, uncertainty-aware state representation that is essential for robustly controlling the robot at high speeds.

\subsection{Flow Matching for High-Speed Policy}
\label{sec:policy_module}

Contemporary generative models, such as Diffusion Policy~\cite{27, 31}, are hindered by a significant latency bottleneck due to their iterative inference process, rendering them unsuitable for high-frequency control tasks~\cite{30}. To circumvent this, we integrate a policy based on \textbf{Flow Matching}~\cite{28, 29}, a technique that enables the generation of an entire action sequence in a single, efficient forward pass. This choice is critical for minimizing the decision-making latency within our perception-control loop.

\begin{figure}
    \centering
    \setlength{\belowcaptionskip}{-0.5cm}
    \includegraphics[width=0.9\linewidth]{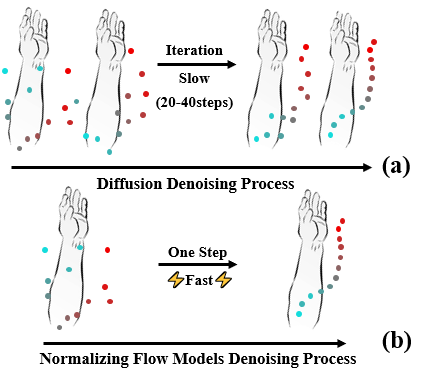}
    \caption{Conceptual comparison of policy inference processes. (a) Diffusion Policies rely on an iterative denoising process, requiring multiple steps to generate an action. (b) Flow Policy enables single-step inference, drastically reducing latency and enabling high-frequency control.}
    \label{fig:inference_comparison}
\end{figure}

The core of this policy is learning to model the trajectory between a simple noise distribution $p_0$ (e.g., a standard Gaussian) and the distribution of expert actions $p_1$. This is achieved by parameterizing a continuous, time-dependent vector field governed by an Ordinary Differential Equation (ODE). Crucially, this vector field is conditioned on the state representation $\mathbf{s}_t$ supplied by our high-frequency perception front-end (Sec.~\ref{sec:perception_module}):
\begin{equation}
    \frac{d\mathbf{x}_t}{dt} = v(\mathbf{x}_t, t | \mathbf{s}_t)
    \label{eq:flow_matching_ode}
\end{equation}
where the neural network $v(\cdot)$ approximates the conditional vector field. Once trained, the policy can directly map a noise vector to a high-quality action sequence in one step, as conceptually illustrated in Fig.~\ref{fig:inference_comparison}.

This single-step inference capability is the cornerstone of our system's real-time performance. It ensures that the policy's minimal latency preserves the temporal advantage gained by our high-frequency perception front-end. The tight coupling of a fast observer with this fast actor actualizes our core design principle of synergy, creating a truly responsive perception-control loop essential for tracking erratic movements in clinical scenarios.

\subsection{Sample-Efficient Sim-to-Real Transfer}
\label{sec:sim2real}

To overcome the reliance on large-scale clinical data, we introduce a three-stage Sim-to-Real training pipeline designed in synergy with our decoupled perception architecture. This co-design leverages the network's structural inductive biases to enable highly sample-efficient adaptation from minimal real-world data.

\paragraph{\textbf{Step 1: Vision Pre-training (Simulation)}}
We first train the decoupled vision front-end on 20,000 simulated image pairs generated from CT volumes. This stage employs an \textbf{enhanced domain randomization} strategy, varying not only visual properties (e.g., brightness) but also crucial US physics parameters (e.g., probe frequency, TGC curves). The objective is to build a robust visual feature foundation that is invariant to both visual and physical domain shifts.

\paragraph{\textbf{Step 2: End-to-End Pre-training (Simulation)}}
Subsequently, the entire framework is trained end-to-end using 1,000 simulated tracking trajectories. This step aims to learn the fundamental visuomotor control logic, enabling the policy to map visual state representations to effective motion plans within the simulated environment.

\paragraph{\textbf{Step 3: Targeted Fine-tuning (Physical Phantom)}}
Finally, to bridge the "reality gap," the pre-trained model is fine-tuned on a minimal dataset of just \textbf{50 expert trajectories}. This stage leverages the inductive biases of our decoupled architecture for maximum sample efficiency. We freeze most weights of the largely domain-invariant \textbf{geometric stream} and concentrate fine-tuning on the domain-sensitive \textbf{semantic stream}. This targeted adaptation facilitates highly efficient domain transfer while preventing catastrophic forgetting of the knowledge acquired in simulation.




\section{EXPERIMENTS}

To validate our central thesis—that a synergistic co-design of perception and control is essential for breaking the latency barrier in dynamic tracking—we conducted a series of experiments to validate our system on both a dynamic phantom and a human volunteer. The experimental protocol was designed to rigorously answer four key questions: (1) How accurately can the system converge? (2) What is the upper limit of its dynamic tracking capability against high-velocity motion? (3) How robust is the system when faced with complex trajectories? (4) How effectively does the system's performance in a in-vivo clinical scenario? Furthermore, to explicitly demonstrate the superiority of our synergistic architecture, we performed targeted ablation studies, quantifying the individual contributions of our framework.

\subsection{Experimental Setup}

\begin{figure}[t]
    \centering
    \setlength{\belowcaptionskip}{-0.5cm}
    \includegraphics[width=0.9\columnwidth]{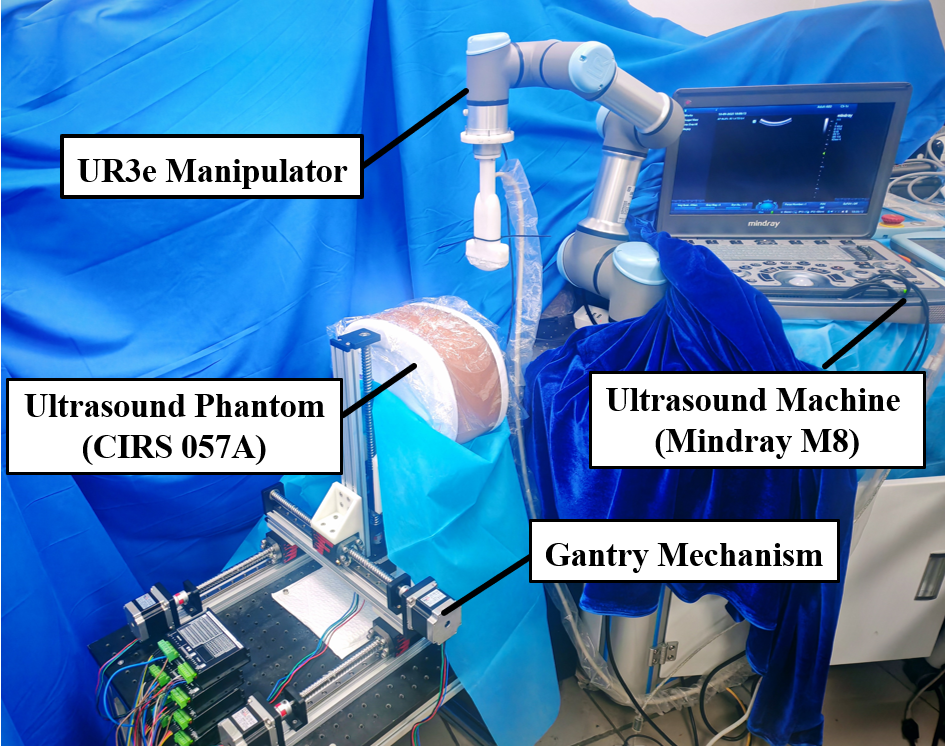}
    \caption{Overview of the experimental setup, showing the UR3e manipulator, the CIRS phantom and the US system.}
    \label{fig:setup}
\end{figure}

Our experimental platform (Fig.~\ref{fig:setup}) comprises a 6-DoF UR3e robotic manipulator, a Mindray M8 US machine with a C5-1s convex probe, and a CIRS Model 057A Abdominal Biopsy Phantom. This phantom was chosen for its clinically relevant anatomical structures (e.g., liver, portal vein), providing a challenging and realistic environment. All perception and control algorithms were executed on a workstation with an NVIDIA RTX 4080 GPU, running Ubuntu 20.04 and ROS Noetic.

\subsection{Baseline Performance Evaluation}

We first evaluate the fundamental performance of our framework in two key scenarios: (1) static and quasi-static repositioning to a target view, and (2) continuous tracking of a target moving at high velocity. These experiments are designed to quantify the system's accuracy, repeatability, and dynamic response capabilities.

\subsubsection{Static and Repositioning Accuracy}
Our framework demonstrates exceptional precision and a wide capture range, achieving a terminal error of approximately \textbf{1.5 mm} in both local and large-scale repositioning tasks. We validated this through two tests: (1) a local convergence test requiring recovery from minor manual displacements, and (2) a large-scale repositioning test where the system had to re-establish the target view after a significant spatial displacement of over $170$~mm.

\begin{figure}[t]
    \centering
    \setlength{\belowcaptionskip}{-0.2cm}
    \includegraphics[width=0.9\columnwidth]{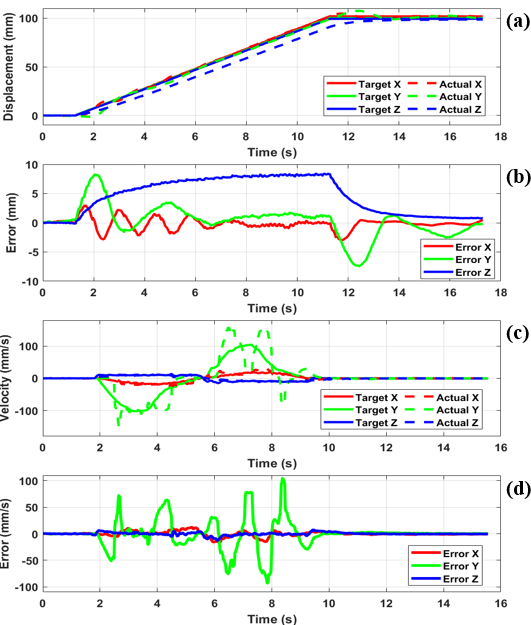}
    \caption{System dynamic performance in large-scale repositioning and high-velocity tracking. \textbf{(a-b)} Convergence dynamics for the $>$170~mm repositioning test. \textbf{(c-d)} Agility during the high-velocity dynamic tracking test.}
    \label{fig:static_perf}
\end{figure}

As quantified in Table~\ref{tab:static_perf}, the system robustly converges with a final positioning error of approximately \textbf{1.52~mm} and near-perfect image similarity (\textbf{NCC$>$0.92}) in both scenarios. The convergence dynamics, visualized for the more demanding large-scale test in Fig.~\ref{fig:static_perf}a and b, confirm a rapid error decay from an total error of over $17$~cm. This robust performance validates the synergistic design of our decoupled perception front-end: the geometric stream effectively handles large geometric deviations, while the semantic stream ensures high-fidelity alignment at the target, enabling the Flow Policy to guide the robot with high precision across vast distances.

\begin{table}[h!]
    \centering
    \caption{Performance in Local Recovery and Global Repositioning Tasks}
    \label{tab:static_perf}
    \renewcommand{\arraystretch}{1.2}
    \resizebox{\columnwidth}{!}{
    \begin{tabular}{lcccc}
        \toprule
        \textbf{Experiment} & \textbf{Movement} & \textbf{Terminal} & \textbf{Terminal} \\
         & \textbf{Dist} (mm) & \textbf{Error} (mm) & \textbf{NCC} \\
        \midrule
        Local Recovery & $21.2$ & $1.5148$ & $0.9481$ \\
        Global Repositioning & $173.2$ & $1.5219$ & $0.9246$ \\
        \bottomrule
    \end{tabular}
    }
\end{table}

\subsubsection{Dynamic Tracking Performance}
\label{sec:dynamic_tracking}
To test end-to-end responsiveness beyond static precision, we designed a high-velocity tracking experiment. This directly stresses the perception-to-action pipeline, where any significant latency—previously a key bottleneck—would result in failure to track the target.

The results provide compelling evidence for the efficacy of our synergistic low-latency design.
As quantified in Table~\ref{tab:dynamic_perf}, the framework successfully tracks a target moving at speeds exceeding \textbf{100~mm/s}, while maintaining a tight mean tracking error of only approximately \textbf{6.12~mm}
The velocity profiles in Fig.~\ref{fig:static_perf}c reveal the policy's agility, showing the actual velocity closely mirroring the high-frequency commands.
More importantly, Fig.~\ref{fig:static_perf}d illustrates that the velocity error remains bounded and low, a direct testament to the system's high control bandwidth.

\begin{figure}[t]
    \centering
    \setlength{\belowcaptionskip}{-0.2cm}
    \includegraphics[width=0.9\columnwidth]{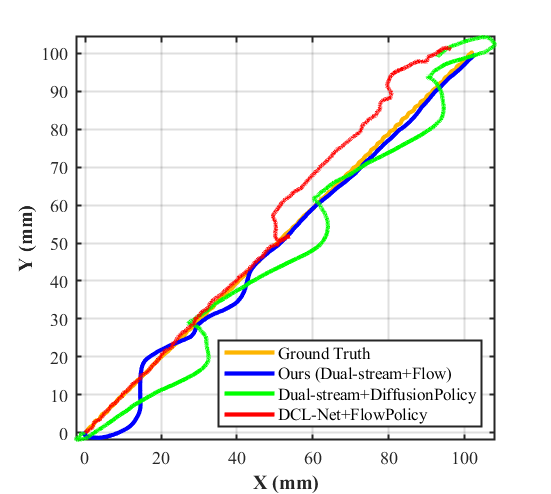}
        \caption{XY-plane trajectory comparison from the dynamic tracking ablation study. Our full framework (blue) closely follows the ground truth (orange). While the Diffusion Policy (green) and DCL-Net module (red) lags significantly.}
    \label{fig:ablation_xy_plot}
\end{figure}

\begin{table}[h!]
    \centering
    \caption{Dynamic Tracking Performance at High Velocity}
    \label{tab:dynamic_perf}
    \renewcommand{\arraystretch}{1.2}
    \begin{tabular}{ccccc}
        \toprule
        \textbf{Max Speed} & \textbf{Avg. Error} & \textbf{Terminal Error} & \textbf{Terminal NCC} \\
        (mm/s) & (mm) & (mm) & &  \\
        \midrule
        102.47 & $6.124 \pm 0.386$ & $1.629$ & $0.9548$ \\
        \bottomrule
    \end{tabular}
\end{table}








\subsection{Robustness on Complex 3D Trajectories}
\label{sec:complex_trajectories}

To assess performance under realistic conditions, we stress-tested the system against 11 complex 3D trajectories designed to probe its limits: spirals for high-curvature tracking, a square wave for abrupt acceleration response, and random paths for stochastic disturbances.

\begin{figure*}[t!] 
    \centering
    \includegraphics[width=0.95\linewidth]{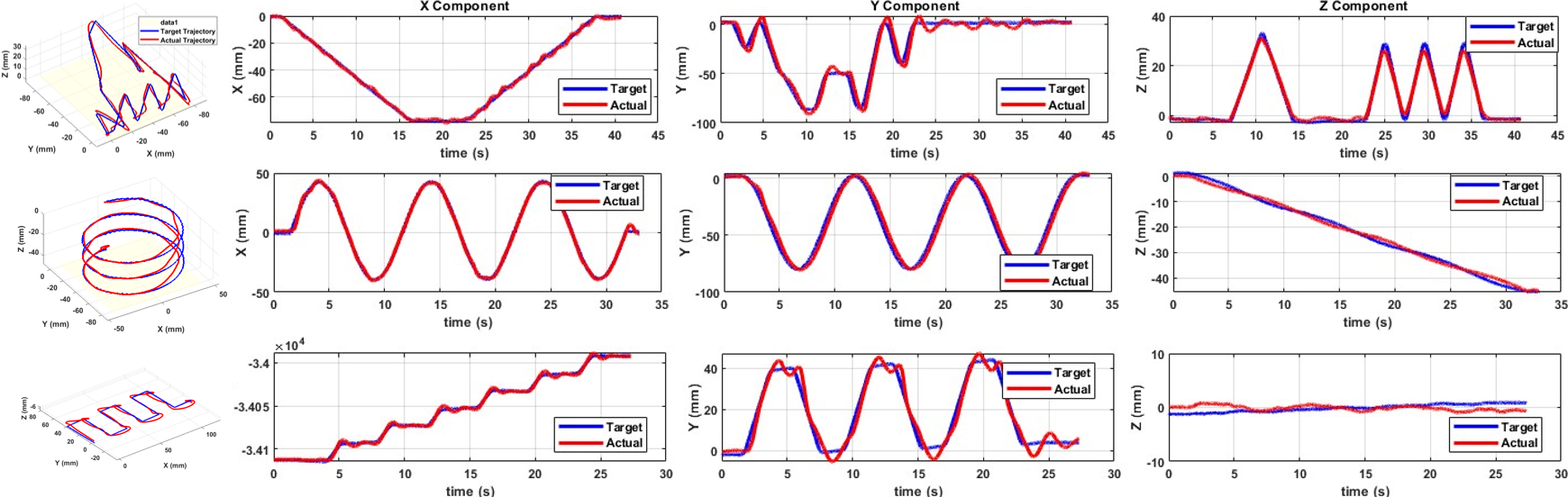}
        \caption{High-fidelity tracking on three complex 3D trajectories (rows, top to bottom: random polyline, spiral, square wave). Each row shows the 3D path (left) and per-axis tracking (X,Y,Z).}
    \label{fig:complex_traj_qualitative}

    \vspace{0.2em} 

    \captionof{table}{Performance on Complex 3D Trajectories}
    \label{tab:complex_traj_table}
    \renewcommand{\arraystretch}{1.2}
    \resizebox{\textwidth}{!}{%
    \begin{tabular}{l c *{5}{c} c *{2}{c}} 
        \toprule
        \textbf{Trajectory Type} & \textbf{Total Time (s)} & \multicolumn{5}{c}{\textbf{Position Metrics (mm)}} & \multicolumn{1}{c}{\textbf{Speed Metrics (mm/s)}} & \multicolumn{2}{c}{\textbf{Image Metrics}} \\
        \cmidrule(lr){3-7} \cmidrule(lr){8-8} \cmidrule(lr){9-10}
        & & \textbf{Error X} & \textbf{Error Y} & \textbf{Error Z} & \textbf{Avg. Tracking Error} & \textbf{Terminal Error} & \textbf{Speed Error} & \textbf{Avg. NCC} & \textbf{Terminal NCC} \\
        \midrule
        \multicolumn{10}{l}{\textit{Spiral-like Trajectories}} \\
        Spiral              & $38.895$ & $1.712 \pm 0.046$ & $3.252 \pm 0.098$ & $1.217 \pm 0.039$ & $4.212 \pm 0.088$ & $1.582$ & $3.759 \pm 0.166$ & $0.9194 \pm 0.0018$ & $0.9592$ \\ 
        Elliptical Spiral   & $32.042$ & $4.800 \pm 0.206$ & $3.232 \pm 0.140$ & $0.912 \pm 0.030$ & $6.313 \pm 0.222$ & $1.139$ & $6.078 \pm 0.269$ & $0.9181 \pm 0.0022$ & $0.9602$ \\ 
        \midrule
        \multicolumn{10}{l}{\textit{Square Wave Trajectory}} \\
        Square Wave         & $30.310$ & $1.392 \pm 0.055$ & $2.836 \pm 0.105$ & $1.536 \pm 0.048$ & $3.906 \pm 0.094$ & $2.959$ & $6.514 \pm 0.254$ & $0.9565 \pm 0.0010$ & $0.9655$ \\ 
        \midrule
        \multicolumn{10}{l}{\textit{Random Polyline Trajectories}} \\
        Random 1            & $42.234$ & $0.900 \pm 0.028$ & $3.740 \pm 0.131$ & $1.081 \pm 0.038$ & $4.337 \pm 0.120$ & $1.686$ & $4.896 \pm 0.207$ & $0.9196 \pm 0.0017$ & $0.9485$ \\ 
        Random 2            & $41.294$ & $0.972 \pm 0.025$ & $2.366 \pm 0.100$ & $1.103 \pm 0.037$ & $3.038 \pm 0.097$ & $2.460$ & $2.915 \pm 0.129$ & $0.9432 \pm 0.0018$ & $0.9572$ \\ 
        Random 3            & $38.177$ & $1.114 \pm 0.046$ & $3.901 \pm 0.140$ & $0.768 \pm 0.033$ & $4.350 \pm 0.138$ & $1.584$ & $5.390 \pm 0.204$ & $0.9432 \pm 0.0016$ & $0.9664$ \\ 
        Random 4            & $39.015$ & $1.349 \pm 0.043$ & $5.716 \pm 0.215$ & $1.241 \pm 0.044$ & $6.357 \pm 0.203$ & $1.684$ & $9.669 \pm 0.351$ & $0.9263 \pm 0.0018$ & $0.9609$ \\ 
        Random 5            & $83.185$ & $1.189 \pm 0.027$ & $4.239 \pm 0.102$ & $0.756 \pm 0.016$ & $4.701 \pm 0.096$ & $1.216$ & $5.981 \pm 0.185$ & $0.9398 \pm 0.0010$ & $0.9555$ \\ 
        Random 6            & $52.897$ & $3.509 \pm 0.065$ & $4.050 \pm 0.121$ & $1.249 \pm 0.035$ & $6.051 \pm 0.105$ & $5.052$ & $7.528 \pm 0.233$ & $0.9204 \pm 0.0013$ & $0.9412$ \\ 
        Random 7            & $55.311$ & $2.066 \pm 0.049$ & $3.948 \pm 0.109$ & $1.064 \pm 0.029$ & $5.019 \pm 0.097$ & $2.719$ & $6.489 \pm 0.201$ & $0.9174 \pm 0.0012$ & $0.9373$ \\ 
        Random 8            & $50.819$ & $2.500 \pm 0.063$ & $3.132 \pm 0.099$ & $1.166 \pm 0.031$ & $4.772 \pm 0.081$ & $4.710$ & $5.868 \pm 0.193$ & $0.9315 \pm 0.0010$ & $0.9403$ \\ 
        \bottomrule
    \end{tabular}%
    }
\end{figure*}

The results (Fig.~\ref{fig:complex_traj_qualitative}, Table~\ref{tab:complex_traj_table}) confirm exceptional robustness. 
Visually, the robot's trajectory (red) tightly follows the ground truth (blue) without overshoot, the system maintained a mean tracking error$<$\textbf{6.4~mm} and, critically, an average \textbf{NCC$>$0.91} across all paths. 
This confirms not just geometric accuracy but the preservation of a stable anatomical view during unpredictable motion. 
Such performance is a direct outcome of our synergistic design, where high-frequency perception provides the timely state updates that enable the predictive policy to master complex dynamics.

\subsection{Ablation Studies and Comparative Analysis}

To validate that our system's performance stems from the synergistic co-design of its components, we conducted ablation studies replacing our modules with strong, SOTA alternatives. For the policy, we selected Diffusion Policy, a dominant paradigm in imitation learning~\cite{27}. For the perception front-end, we chose DCL-Net, a leading framework specifically designed for dynamic US registration~\cite{DCL-Net}. We evaluated these variants on the high-velocity dynamic tracking task to expose critical latency bottlenecks.

The results, summarized in Table~\ref{tab:ablation} and visualized in Fig.~\ref{fig:ablation_xy_plot}, provide conclusive evidence for our thesis.

First, the Diffusion Policy variant revealed a severe latency bottleneck. Its inference time of over 128~ms corresponds to a sluggish control frequency of only \textbf{8~Hz}, fundamentally limiting its dynamic response. This latency not only capped its tracking speed at \textbf{31~mm/s} but also resulting in a large repositioning error of \textbf{9.84~mm}.

Second, and more critically, the system with the DCL-Net front-end \textbf{failed to converge} in dynamic tracking experiment. While the DCL-Net module itself is computationally fast (14~ms), its architecture is not co-designed to provide the stable state estimates required by the policy, leading to immediate instability.

In stark contrast, our framework achieves a low \textbf{16.2 ms} latency, enabling a \textbf{62~Hz} control loop. This high bandwidth is the prerequisite for both its superior dynamic tracking ($>$\textbf{100~mm/s}) and its precise final convergence ($<$\textbf{1.6~mm} error). These results demonstrate that high-performance robotic US is an emergent property that arises only from the tight, synergistic integration of perception and control.

\begin{table}[h!]
    \centering
    \caption{Ablation and Comparative on Dynamic Tracking}
    \label{tab:ablation}
    \renewcommand{\arraystretch}{1.3}
    \resizebox{\columnwidth}{!}{%
    \begin{tabular}{l | c c | c c}
        \toprule
        \multirow{2}{*}{\textbf{Framework Configuration}} & \multicolumn{2}{c|}{\textbf{Latency Metrics}} & \multicolumn{2}{c}{\textbf{Performance Metrics}} \\
        \cmidrule(lr){2-3} \cmidrule(lr){4-5}
        & \textbf{Time (ms)} & \textbf{Freq. (Hz)} & \textbf{Max. Speed (mm/s)} & \textbf{Error (mm)} \\
        \midrule
        \textbf{Ours (Dual-stream + Flow)} & $\approx 16.2$ & $\approx 62$ & \textbf{102.47} & \textbf{1.52} \\
        Dual-stream + Diffusion & $\approx 128.2$ & $\approx 8$ & 30.98 & 9.84 \\
        DCL-Net + Flow & $\approx 13.7$ &  73 & \textbf{Failed to Converge} & 6.508 \\
        \bottomrule
    \end{tabular}%
    }
\end{table}








\subsection{Robustness to Out-of-Plane Rotational Disturbances}
\label{sec:rotation_robustness}

Our framework is designed for 3D translational servoing. To characterize its operational limits, we evaluated its stability under unmodeled Z axis rotational disturbances. We focus on this Z axis rotation as other rotations (tilting around the X and Y axes) may reduce US imaging quality and is considered out of the scope of this work~\cite{my_robio}. In a spiral tracking experiment, we introduced rotational offsets from 0\textdegree~to 25\textdegree.

As summarized in Table~\ref{tab:rotation_robustness}, the results clearly define the system's robustness boundary. 
The framework remains stable up to a \textbf{15\textdegree}~offset, maintaining high image similarity and minimal positional error. 
Beyond this threshold, performance degrades rapidly, with control instability at 20\textdegree~and tracking failure at 25\textdegree.
This failure mode stems directly from our perception front-end's inductive bias: the network is trained to interpret visual shearing as \textit{translational} motion. 
The resulting instability stems from a predictable perceptual ambiguity—the network mistaking rotational shearing for translation—not from simple tracking drift.

\begin{table}[h!]
    \centering
    \caption{Stability under Out-of-Plane Rotational Error}
    \label{tab:rotation_robustness}
    \renewcommand{\arraystretch}{1.3}
    \begin{tabular}{cccc}
        \toprule
        \textbf{Rotational Offset} & \textbf{Positional Error} & \textbf{Avg. NCC} & \textbf{Status} \\
        \midrule
        0\textdegree  & $4.212 \pm 0.088$ & $0.9181 \pm 0.0018$ & Stable \\
        5\textdegree  & $4.659 \pm 0.093$ & $0.9161 \pm 0.0015$ & Stable \\
        10\textdegree & $5.096 \pm 0.103$ & $0.9477 \pm 0.0014$ & Stable \\
        15\textdegree & $8.170 \pm 0.179$ & $0.9370 \pm 0.0007$ & Stable \\
        20\textdegree & $9.916 \pm 0.183$ & $0.9153 \pm 0.0013$ & \textbf{Unstable} \\
        25\textdegree & N/A & $0.6962 \pm 0.0096$ & \textbf{Failed} \\
        \bottomrule
    \end{tabular}
\end{table}






\subsection{In-vivo Validation on Human Volunteers}
To validate clinical applicability, we conducted an in-vivo study where the framework autonomously tracked over \textbf{20~cm} along a human volunteer's forearm. Despite challenges from non-rigid tissue and physiological motion, the system successfully completed the \textbf{27-second} scanning. A high terminal image similarity (\textbf{NCC of 0.946}) was reached, confirming the framework's robustness in a realistic environment and bridging the gap from phantom to clinical application.

\begin{figure}[t]
    \centering
    \setlength{\belowcaptionskip}{-0.5cm}
    \includegraphics[width=0.9\columnwidth]{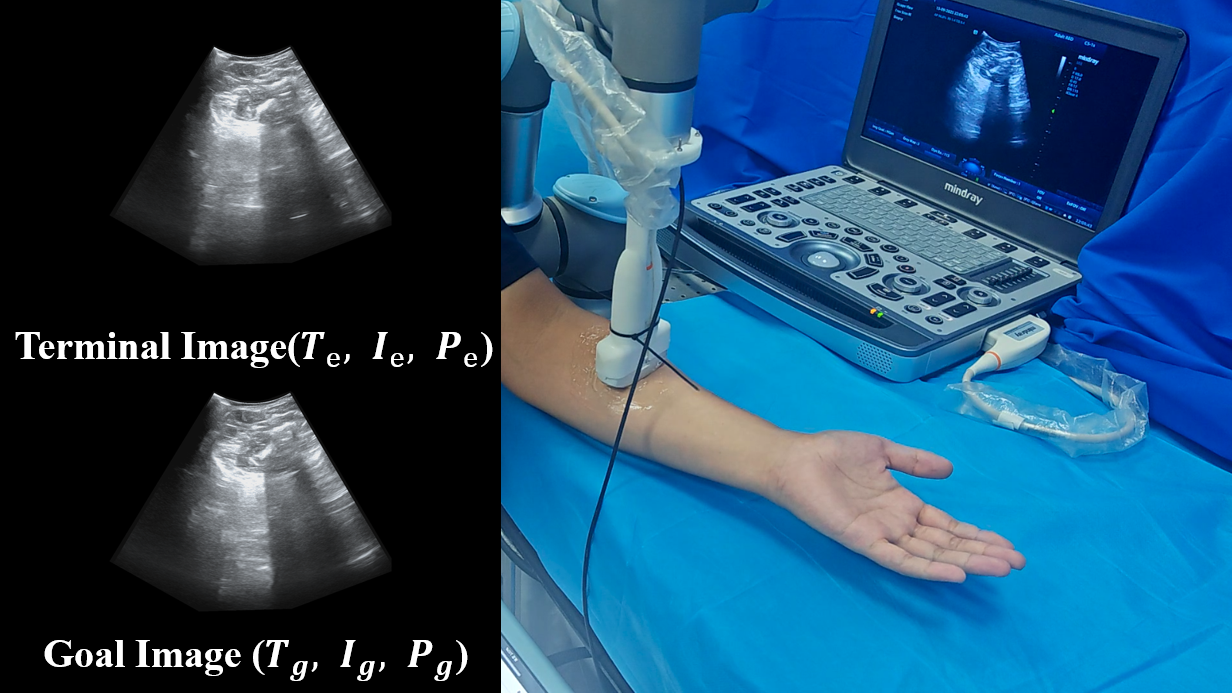}
        \caption{In-vivo validation of the proposed framework on a human volunteer.}
    \label{fig:human_experiment}
\end{figure}

\section{Conclusion}

In this work, we broke the latency barrier in dynamic robotic US tracking. We presented a framework founded on the principle of synergistic co-design, which synergizes a high-frequency, decoupled perception front-end with a single-step Flow Matching policy. Experiments on a physical platform, including a dynamic phantom and a human volunteer, validate our approach, demonstrating a closed-loop frequency exceeding 60~Hz, successful tracking of complex trajectories at over 100~mm/s, and sample-efficient Sim-to-Real transfer. While the current system focuses on 3D translation, our immediate future work will focus on extending the system to full 6-DoF pose control and validating its clinical potential through more comprehensive human studies across various anatomical regions.


\bibliographystyle{IEEEtran}
\bibliography{IEEEabrv,xbib}

\begin{thebibliography}{10}
\providecommand{\url}[1]{#1}
\csname url@rmstyle\endcsname
\providecommand{\newblock}{\relax}
\providecommand{\bibinfo}[2]{#2}
\providecommand\BIBentrySTDinterwordspacing{\spaceskip=0pt\relax}
\providecommand\BIBentryALTinterwordstretchfactor{4}
\providecommand\BIBentryALTinterwordspacing{\spaceskip=\fontdimen2\font plus
\BIBentryALTinterwordstretchfactor\fontdimen3\font minus
  \fontdimen4\font\relax}
\providecommand\BIBforeignlanguage[2]{{%
\expandafter\ifx\csname l@#1\endcsname\relax
\typeout{** WARNING: IEEEtran.bst: No hyphenation pattern has been}%
\typeout{** loaded for the language `#1'. Using the pattern for}%
\typeout{** the default language instead.}%
\else
\language=\csname l@#1\endcsname
\fi
#2}}

\bibitem{8}
Z.~Jiang, S.~E. Salcudean, and N.~Navab, ``Robotic ultrasound imaging:
  State-of-the-art and future perspectives,'' \emph{Medical image analysis},
  vol.~89, p. 102878, 2023.

\bibitem{9}
K.~Munir, A.~F. Al-Battal, A.~Al-Sheghri, H.~Becher, M.~Noga, and
  K.~Punithakumar, ``A survey of autonomous robotic ultrasound scanning
  systems,'' \emph{IEEE Access}, 2025.

\bibitem{2}
Z.~Jiang, N.~Danis, Y.~Bi, M.~Zhou, M.~Kroenke, T.~Wendler, and N.~Navab,
  ``Precise repositioning of robotic ultrasound: Improving registration-based
  motion compensation using ultrasound confidence optimization,'' \emph{IEEE
  Transactions on Instrumentation and Measurement}, vol.~71, pp. 1--11, 2022.

\bibitem{10}
Z.~Jiang, Z.~Li, M.~Grimm, M.~Zhou, M.~Esposito, W.~Wein, W.~Stechele,
  T.~Wendler, and N.~Navab, ``Autonomous robotic screening of tubular
  structures based only on real-time ultrasound imaging feedback,'' \emph{IEEE
  Transactions on Industrial Electronics}, vol.~69, no.~7, pp. 7064--7075,
  2022.

\bibitem{4}
X.~Ma, M.~Zeng, J.~C. Hill, B.~Hoffmann, Z.~Zhang, and H.~K. Zhang, ``Guiding
  the last centimeter: Novel anatomy-aware probe servoing for standardized
  imaging plane navigation in robotic lung ultrasound,'' \emph{IEEE
  Transactions on Automation Science and Engineering}, 2024.

\bibitem{30}
Q.~Rouxel, A.~Ferrari, S.~Ivaldi, and J.-B. Mouret, ``Flow matching imitation
  learning for multi-support manipulation,'' in \emph{2024 IEEE-RAS 23rd
  International Conference on Humanoid Robots (Humanoids)}.\hskip 1em plus
  0.5em minus 0.4em\relax IEEE, 2024, pp. 528--535.

\bibitem{20}
J.~Zhou, H.~Tian, W.~Wang, \emph{et~al.}, ``Fully automated thyroid ultrasound
  screening utilizing multi-modality image and anatomical prior,''
  \emph{Biomedical Signal Processing and Control}, vol.~87, p. 105430, 2024.

\bibitem{17}
Z.~Jiang, Y.~Bi, M.~Zhou, Y.~Hu, M.~Burke, and N.~Navab, ``Intelligent robotic
  sonographer: Mutual information-based disentangled reward learning from few
  demonstrations,'' \emph{The International Journal of Robotics Research},
  vol.~43, no.~7, pp. 981--1002, 2024.

\bibitem{19}
Y.~Huang, W.~Xiao, C.~Wang, H.~Liu, R.~Huang, and Z.~Sun, ``Towards fully
  autonomous ultrasound scanning robot with imitation learning based on
  clinical protocols,'' \emph{IEEE Robotics and Automation Letters}, vol.~6,
  no.~2, pp. 3671--3678, 2021.

\bibitem{5}
S.~Ipsen, D.~Wulff, I.~Kuhlemann, A.~Schweikard, and F.~Ernst, ``Towards
  automated ultrasound imaging—robotic image acquisition in liver and
  prostate for long-term motion monitoring,'' \emph{Physics in Medicine \&
  Biology}, vol.~66, no.~9, p. 094002, 2021.

\bibitem{3}
T.~Chen, X.~Zhao, Y.~Zhang, G.~Zheng, L.~Hou, Q.~Ling, B.~Tao, and Z.~Yin,
  ``Ultrasound-guided robotic autonomous operation based on real-time
  deformation tracking and prediction,'' \emph{IEEE Transactions on Industrial
  Informatics}, 2024.

\bibitem{1}
J.~Tan, J.~Li, Y.~Li, B.~Li, Y.~Leng, Y.~Rong, and C.~Fu, ``Autonomous
  trajectory planning for ultrasound-guided real-time tracking of suspicious
  breast tumor targets,'' \emph{IEEE Transactions on Automation Science and
  Engineering}, vol.~21, no.~3, pp. 2478--2493, 2023.

\bibitem{ning1}
G.~Ning, H.~Liang, X.~Zhang, and H.~Liao, ``Autonomous robotic ultrasound
  vascular imaging system with decoupled control strategy for
  external-vision-free environments,'' \emph{IEEE Transactions on Biomedical
  Engineering}, vol.~70, no.~11, pp. 3166--3177, 2023.

\bibitem{43}
G.~Ning, H.~Liang, X.~Zhang, and H.~Liao,
  ``Inverse-reinforcement-learning-based robotic ultrasound active compliance
  control in uncertain environments,'' \emph{IEEE Transactions on Industrial
  Electronics}, vol.~71, no.~2, pp. 1686--1696, 2024.

\bibitem{27}
C.~Chi, Z.~Xu, S.~Feng, E.~Cousineau, Y.~Du, B.~Burchfiel, R.~Tedrake, and
  S.~Song, ``Diffusion policy: Visuomotor policy learning via action
  diffusion,'' \emph{The International Journal of Robotics Research}, p.
  02783649241273668, 2023.

\bibitem{31}
H.~Wang, Y.~Long, Y.~Chen, H.-C. Yip, M.~Scheppach, P.~W.-Y. Chiu, Y.~Yam,
  H.~M.-L. Meng, and Q.~Dou, ``Learning dissection trajectories from expert
  surgical videos via imitation learning with equivariant diffusion,''
  \emph{Medical Image Analysis}, p. 103599, 2025.

\bibitem{28}
Y.~Fang, X.~Zhang, H.~Cheng, X.~Zang, R.~Song, and J.~Zhao, ``Flow policy:
  Generalizable visuomotor policy learning via flow matching,'' \emph{IEEE/ASME
  Transactions on Mechatronics}, 2025.

\bibitem{29}
Q.~Zhang, Z.~Liu, H.~Fan, G.~Liu, B.~Zeng, and S.~Liu, ``Flowpolicy: Enabling
  fast and robust 3d flow-based policy via consistency flow matching for robot
  manipulation,'' in \emph{Proceedings of the AAAI Conference on Artificial
  Intelligence}, vol.~39, no.~14, 2025, pp. 14\,754--14\,762.

\bibitem{24}
Y.~Long, A.~Lin, D.~H.~C. Kwok, L.~Zhang, Z.~Yang, K.~Shi, L.~Song, J.~Fu,
  H.~Lin, W.~Wei, \emph{et~al.}, ``Surgical embodied intelligence for
  generalized task autonomy in laparoscopic robot-assisted surgery,''
  \emph{Science Robotics}, vol.~10, no. 104, p. eadt3093, 2025.

\bibitem{13}
A.~Tyagi, A.~Tyagi, M.~Kaur, R.~Aggarwal, K.~D. Soni, J.~Sivaswamy, and
  A.~Trikha, ``Nerve block target localization and needle guidance for
  autonomous robotic ultrasound guided regional anesthesia,'' in \emph{2024
  IEEE/RSJ International Conference on Intelligent Robots and Systems
  (IROS)}.\hskip 1em plus 0.5em minus 0.4em\relax IEEE, 2024, pp. 5867--5872.

\bibitem{38}
D.~Dall’Alba, L.~Busellato, T.~R. Savarimuthu, Z.~Cheng, and I.~Iturrate,
  ``Imitation learning of compression pattern in robotic assisted ultrasound
  examination using kernelized movement primitives,'' \emph{IEEE Transactions
  on Medical Robotics and Bionics}, 2024.

\bibitem{6}
X.~Liu, C.~He, M.~Wu, A.~Ping, A.~Zavodni, N.~Matsuura, and E.~Diller,
  ``Transformer-based robotic ultrasound 3d tracking for capsule robot in gi
  tract,'' \emph{International Journal of Computer Assisted Radiology and
  Surgery}, pp. 1--8, 2025.

\bibitem{11}
E.~Zakeri, A.~Spilkin, H.~Elmekki, A.~Zanuttini, L.~Kadem, J.~Bentahar, W.-F.
  Xie, and P.~Pibarot, ``Robust deep feature ultrasound image-based visual
  servoing: focus on cardiac examination,'' \emph{IEEE/ASME Transactions on
  Mechatronics}, 2025.

\bibitem{41}
H.~Yoon and S.-W. Kim, ``Efficient and robust fabrication of soft sensors via
  injection with auxiliary suction in multilayered microchannels with a liquid
  metal alloy,'' \emph{IEEE Sensors Journal}, vol.~25, no.~13, pp.
  23\,948--23\,957, 2025.

\bibitem{DCL-Net}
H.~Guo, S.~Xu, B.~Wood, and P.~Yan, ``Sensorless freehand 3d ultrasound
  reconstruction via deep contextual learning,'' in \emph{International
  Conference on Medical Image Computing and Computer-Assisted
  Intervention}.\hskip 1em plus 0.5em minus 0.4em\relax Springer, 2020, pp.
  463--472.

\bibitem{my_robio}
Y.~Qian, Y.~Zhang, M.~Q.-H. Meng, and L.~Liu, ``Autonomous in-plane normal
  positioning in robotic ultrasound scanning,'' in \emph{2024 IEEE
  International Conference on Robotics and Biomimetics (ROBIO)}, 2024, pp.
  342--347.

\end{thebibliography}

\end{document}